\documentclass{article}
\usepackage{spconf,amsmath,epsfig}
\usepackage{color}
\usepackage{url}
\usepackage{graphicx}
\usepackage{epstopdf}
\usepackage{epsfig}
\usepackage{booktabs}
\usepackage{blindtext}
\usepackage{setspace}
\usepackage{float}

\newcommand{\myboldbullet}[1]{\vspace{0.2cm} \noindent $\bullet$ {\bf #1}}

\newlength{\bibitemsep}
\newlength{\bibparskip}
\let\oldthebibliography\thebibliography
\renewcommand\thebibliography[1]{%
  \oldthebibliography{#1}%
  \setlength{\parskip}{\bibitemsep}%
  \setlength{\itemsep}{\bibparskip}%
}
\expandafter\def\expandafter\normalsize\expandafter{%
    \normalsize
    \setlength\abovedisplayskip{5pt}
    \setlength\belowdisplayskip{10pt}
    \setlength\abovedisplayshortskip{10pt}
    \setlength\belowdisplayshortskip{10pt}
}

\begin{document}\sloppy

\def\x{{\mathbf x}}
\def\L{{\cal L}}

\title{Utilizing High-level Visual Feature for \\ Indoor Shopping Mall Navigation}
%

\name{Ziwei Xu \textsuperscript{1*} \thanks{* These authors contributed equally to this work.}, Haitian Zheng  \textsuperscript{2*} \footnotemark[1], Minjian Pang  \textsuperscript{2}, Yangchun Zhu \textsuperscript{1}, Xiongfei Su  \textsuperscript{2}, Guyue Zhou  \textsuperscript{3}, Lu Fang  \textsuperscript{2}}

\address{University of Science and Technology of China \textsuperscript{1}\\ Hong Kong University of Science and Technology \textsuperscript{2}\\ Dajiang Innovations Co., Ltd. \textsuperscript{3}}

\maketitle

\begin{abstract}
Towards robust and convenient indoor shopping mall navigation, we propose a novel learning-based scheme to utilize the high-level visual information from the storefront images captured by personal devices of users. Specifically, we decompose the visual navigation problem into localization and map generation respectively. Given a storefront input image, a novel feature fusion scheme (denoted as FusionNet) is proposed by fusing the distinguishing DNN-based appearance feature and text feature for robust recognition of store brands, which serves for accurate localization. Regarding the map generation, we convert the user-captured indicator map of the shopping mall into a topological map by parsing the stores and their connectivity. Experimental results conducted on the real shopping malls demonstrate that the proposed system achieves robust localization and precise map generation, enabling accurate navigation.
\end{abstract}
\section{Introduction}
In absence of a portable, low-cost positioning system like GPS for outdoor localization, indoor positioning system (IPS) has been a long-lasting and attractive research topic. Infrastructure-based indoor positioning systems which make use of pre-installed infrastructures such as RFID \cite{reza2009investigation}, configured fluorescent lights \cite{XLiu:10} or Wi-Fi access points \cite{NChang:10} have achieved impressive performance in real scene. The infrastructure-free IPS, on the contrary, superior and challenging by itself, attracts a lot of attention nowadays. Resorting to image retrieval techniques, \cite{huitl2012virtual} \cite{liang2013image} \cite{guan2016vision} proposed vision-based IPS which can tell a user's position using photos taken by smart-phones. However, all of these methods require an off-line database building process which is time-consuming and costly. 

Recent advances on robotics and computer vision throw light on IPS. Simultaneous Localization and Mapping (SLAM) techniques and Visual Odometry (VO) which can accurately estimate motion make these methods promising for IPS. \cite{mur2015orb} proposed a monocular SLAM system which uses bag-of-words for place recognition \cite{galvez2012bags}. \cite{forster2014svo} proposed an accurate monocular VO algorithm which can run at 55 FPS on an ARM platform. \cite{zhangcompact} proposed a real-time indoor and outdoor localization system with visual lidar odometry and mapping \cite{zhang2015visual} and ranked highest on Microsoft Indoor Localization Competition in 2016 \cite{ipsn2016}. However, running SLAM or VO in practice means users have to record videos by cameras or laser transceivers.

To address this issue, Wang et al. \cite{SWang:1} proposed an approach which relies on text recognition \cite{Goodfellow:13} \cite{Jaderberg:14c} for shop candidate classification. Specifically, shops in an image are classified by text recognition and serve as landmarks for coarse-level localization (i.e., localization by shop classification). This approach is extendable and flexible, since it does not require large amount of pre-captured data about the indoor scenes except a pre-labelled floor plan. 

\begin{figure*}[t]
	\centering
	\includegraphics[width=.9\textwidth]{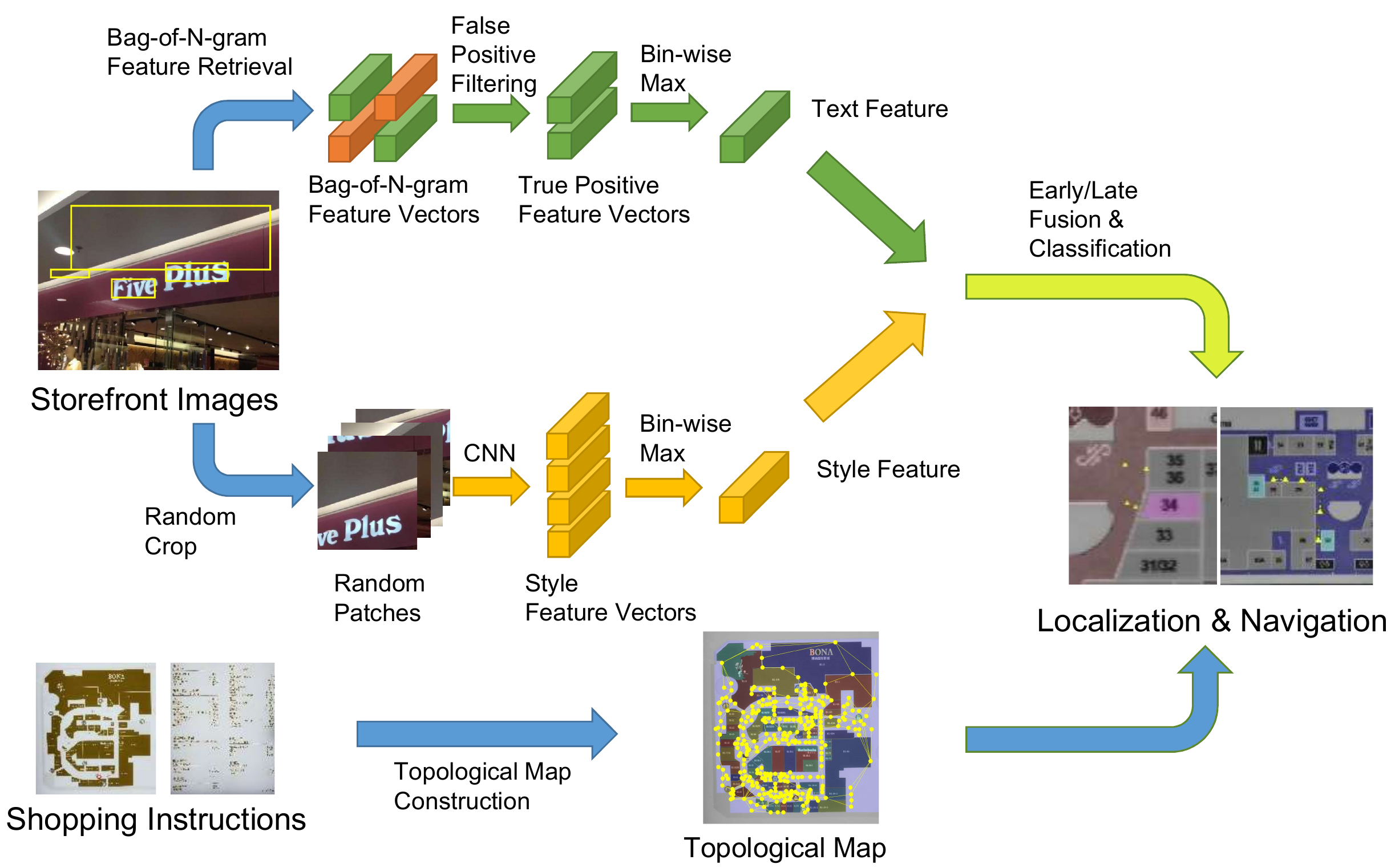}
	\caption{System Architecture. Style features and text features are extracted from storefront images and fused for shop classification. Images of shopping instructions are parsed and a topological map of the shopping mall is constructed. Localization and navigation is performed based on the classification result and the topological map.}
	\label{sysArch}
\end{figure*}

Inspired by the idea of localization by shop classification, we propose a flexible and robust indoor shopping mall navigation system.
Three major contributions are made in this paper. Firstly, we propose a novel style feature and fuse this feature with text feature for robust classification. We prove by experiment that our fusion framework, the FusionNet, outperforms \cite{SWang:1}. Moreover, we build a storefront image dataset and plan to make it publicly available. Secondly, we introduce a shopping instruction parsing method which can automatically build topological map from photos taken by smart-phones. Thirdly, with robust shop recognition and automatic topological map construction, we put forward a flexible indoor navigation system which successfully works in real scenes. The architecture of our system is shown in Fig. \ref{sysArch}.

\section{Localization by Classification}
\label{section-store-recog}
Localization by classification is the most common way that shoppers adopt when they get lost in a shopping mall. Recognizing a shop and matching it with a map can produce an instant estimation of location. In this section, we introduce our feature fusion method for shop classification.


\subsection{Data Collection}
We collected storefront images, from Adidas to Zippo, of 56 different classes of shop from the Internet. The collected images include varieties of style, decoration, and tone of color. For most classes, up to 88 images were collected, and for some rarely seen shop brands, there were still at least 20 images available. We collected a total of 2,876 images, with 51 images collected for each brand on average. Within each class, images were divided into training set and test set, with a ratio of 4:1. Interested readers are referred to the supplemental material for a detailed description of the dataset.

\subsection{Style Features for Classification}
Many famous brands decorate their shops in a particular way so that customers can easily recognize and remember their storefronts. For example, Adidas stores usually have a black background decorated by unique black-white stripes, while the common decoration for Gucci shops is a golden color with a grid-patterned window. These visual patterns are distinctive for different brands, and are usually stable among shopping malls.  

To utilize these visual patterns for shop recognition, we adopt transfer learning to learn discriminative visual representations. Specifically, random patches along with their shop brand ground truths are fetched from our dataset to fine-tune the AlexNet \cite{Krizhevsky:12}. The 4096-dimensional feature of the trained network (i.e., the output of the 7th layer) is used to represent shop styles.

In the testing phase, we randomly choose 16 patches in an image and calculate each patch's feature vector. After this, a bin-wise max operation (we find that features obtained with bin-wise max operation outperform features that obtained with bin-wise average operation) is performed on each of the feature dimensions. A 4096-dimensional vector is thereby generated to represent style feature.

\subsection{Improving Text Features}
\label{subsect:text_feature}
The previous work \cite{SWang:1} made it feasible to utilize text detection and bag-of-N-grams feature  for shop recognition. However, from experiments with our dataset, issues such as false positive detection and irrelevant text made text features unreliable.

We addressed this problem by training a linear classifier to predict the reliability of text detection. Specially, for each text detection result, the corresponding 10000-dimensional text feature plus 4 geometrical features -- ${w}$, ${h}$, ${w+h}$ and ${w/h}$, which is the width, height, scale and shape of the bounding box -- are used as input. A logistic regression classifier is trained to reject unreliable text detection. In practice, the simple linear classifier filtered out most of the false positives.

Another modification on text-based classification is made in this paper. \cite{SWang:1} designed the \emph{n-gram score} as a linear summation of bag-of-N-grams feature with predefined 0-or-1 weights;
the larger n-gram score means the higher likelihood of the text belonging to a class. In this paper, the n-gram score is generalized to a linear classifier with learned weights, and we use classification score to predict the class type. Intuitively, the learned weight would avoid putting large weights on grams like single letters (e.g., ``a'', ``b'', etc.) or the common combinations (e.g., ``ti'', ``la'', which occur in many brands), thus should be more discriminative. 

To handle the problem of overfitting, we reduce the feature dimensionalities by truncating feature vectors and keeping components which correspond to n-grams with a length less than or equal to $k$. The truncated feature is denoted as ${\mathcal{G}_k}$ and we will refer to $k$ as the \textbf{order of the text feature}. In experiments, we found that ${\mathcal{G}_2}$ performs the best on our dataset (please refer to Section \ref{exp-store-recog} for details).

\subsection{Fusion of Style and Text}
Style and text features describe very different properties of a storefront respectively. Therefore, fusing the two features for classification should improve classification accuracy.

The most straightforward way for fusion is to concatenate the two feature vectors into a new vector and then perform a logistic regression on it. The linear regression is a natural scheme because the last layer of the fine-tuned style CNN can be seen as a linear classifier, and also because of the linear extension we made on the text score in Section \ref{subsect:text_feature}. We will refer to this model as the early FusionNet (\textbf{E-FusionNet}) model. 

As shown by \cite{Snoek:05}, the early fusion model usually performs worse than the late fusion model. Based on that observation, we also tested the late fusion. Specifically, we get the normalized text score and normalized style score separately from two classifiers and add the two scores together as the final class score. The late fusion model can be expressed as

\begin{equation}
	\label{eq-additive-model}
	\vec{y} = (1-\alpha) f_{a}(W_t \vec{x_t}) + \alpha f_{a}(W_s \vec{x_s}),
\end{equation}
where $\vec{y}$ is the class score, $\vec{x_t} \in \{\mathcal{G}_k\ | k=1,2,3,...\}$ is the text feature vector, $\vec{x_s}$ is the style feature, $W_t$ and $W_s$ are linear classifier weights learned separately from text features and style features, $f_{a}$ is the sigmoid activation function, and $\alpha \in [0,1]$ is a tunable parameter that controls the weights of the two scores. The selection of $\alpha$ is discussed in Section \ref{section-store-recog}. We will refer to model in equation \ref{eq-additive-model} as the \textbf{L-FusionNet} model.

\section{Topological Map Construction}
\label{section-tp-map-constr}
Shopping mall operators usually provide shoppers with sufficient topological information about the shopping mall in the form of shopping instructions. A typical set of shopping instructions consists of an indicator map and a list of shops. In this section, we introduce our method of building a topological map from shopping instructions. 

\subsection{Text Detection and Recognition}
In most shopping mall indicator maps, there is a significant contrast between text elements and other components. Therefore, text can be extracted by detecting maximally stable extremal regions (MSERs) \cite{obdrvzalek2009detecting}. We detect MSERs in the image as connected components. Detected connected components are filtered based on their size, eccentricity and aspect ratio (i.e., ${w/h}$, where $w$ is the width of the bounding box and $h$ is the height). Filtered connected components are clustered using the run length smoothing algorithm (RLSA) \cite{Nikolaou:2010} and then recognized by the open source tesseract-ocr software package \cite{RSmith}. After being localized and recognized,text and icons are removed using inpainting method introduced in \cite{Garcia:2010} and \cite{wang2012three}. 

\subsection{Road and Shop Segmentation}
We apply the Statistical Region Merging (SRM) \cite{Nock:2004} for map segmentation. This segmentation method can be tuned by a Q value, which indicates the approximate number of expected segments. We are particularly interested in the road component on which our topological map is built; therefore, we perform an initial SRM on the whole map with a small Q (16 in our experiment). This gives a coarse output where road component, shop blocks and other components, like the background, are separated. To identify the road component, we calculate a road score for all components. The score is defined as
\begin{equation}
	\label{segAA}
	s_i = \frac{Coverage(c_i)}{Area(c_i)}-H(c_i)-Deviation(c_i).
\end{equation}
Here, $c_i$ denotes the $i^{th}$ component, $H(c_i)$ is the number of holes inside $c_i$, and $Area(\cdot)$ is the area of $c_i$'s circumscribed rectangle. $Deviation(c_i)$ is the distance between $c_i$'s centroid and the center of the whole image. $Coverage(c_i)$ is defined as
\begin{equation}
	Coverage(c_i) = \sum_{j \neq i}  I(c_i \cap c_j),
\end{equation}
where $I(\cdot)$ is the indicator function. All these values are normalized into the same scale. The component with the largest road score is considered as the road component.

The rest of connected components are shop blocks. A second SRM is performed on the indicator map with a larger Q value (512 in our experiment) to separate these shop blocks. This operation is confined within the area wrapped by the road component's circumscribed rectangle so that items on the background will not interfere with the subsequent road segmentation process.

The road component is further segmented for navigation. We treat each pixel of the road component as an observation spot. For each observation spot, the algorithm searches its neighborhood for shop blocks and store the ID of such blocks as landmarks of the spot. Pixels (spots) with same landmarks are grouped together to form a node.

\subsection{Shop List Parsing}
When wandering in a shopping mall, a shopper is interested in the name of a shop rather than its ID number on the indicator map. Therefore, it is important to map shop IDs with shop names by parsing the shop lists.

A shop list is cut into several blocks using the Recursive XY-cut algorithm. Each of the blocks is a column of shop names, shop IDs or a mixture of both. We then split or merge segments so that each segment contains only one column of shop names and one column of shop ID numbers. Such segments are split into different lines where each line is a mixture of a shop name and a shop ID. A name-ID map is finally constructed between the name and the ID on such lines. 

A topological map is constructed based on the topological and semantic information extracted in all steps above. An undirected weighted graph is constructed to represent the topological layout of the shopping mall. Each node has a landmark array storing IDs of all shop blocks in its neighborhood. The weight of an edge is the distance between the centroids of two nodes.

\section{Experiments}
\label{section-expr}

\begin{figure}[t]
	\centering
	\includegraphics[width=0.49\textwidth]{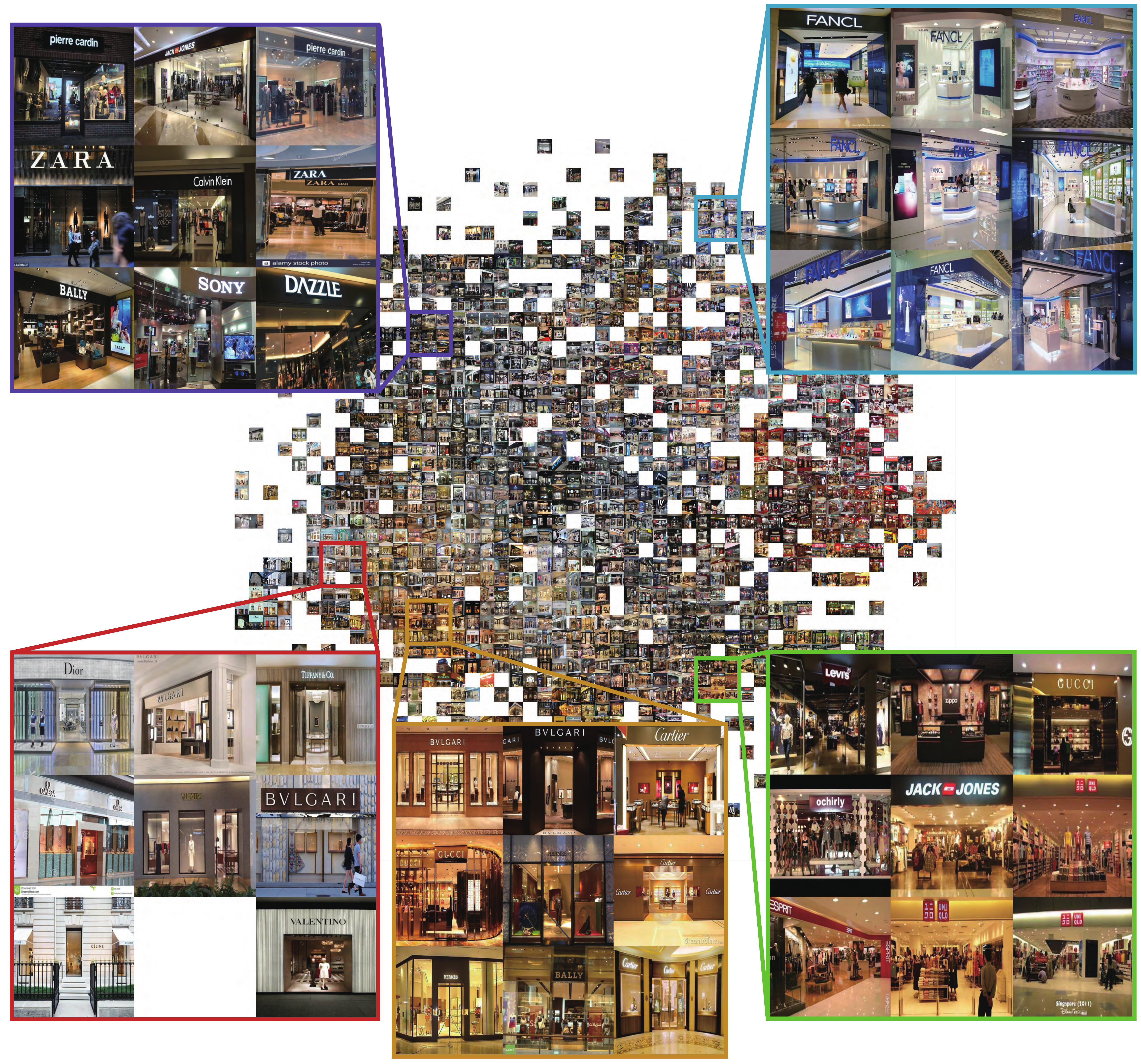}
	\caption{2-D embedding visualization of style feature using t-SNE \cite{tSNE}. Five groups of clustered images are shown in the colored boxes. The supplementary material includes a high-resolution version of this visualization.}
	\label{fig:style-feature-visualize}
\end{figure}

\subsection{Storefront Recognition}
\label{exp-store-recog}
The accuracy for shop classification is the metric we are interested in. In our experiment, we compared our feature fusion models with style-only models and text-only models. Moreover, we compared our method with \cite{SWang:1}. The result is shown in Table \ref{table-performance-compare}.

\myboldbullet{Style Features}
For style feature retrieval, we fine-tuned the AlexNet with our dataset. Our implementation is based on the deep learning framework Caffe \cite{jia2014caffe}. After 100,000 iterations of SGD optimization, with a learning rate of 1e-7 for all convolution layers and 1e-3 for all fully connected layers, the test accuracy reached 44.67\%. This classification accuracy on individual patches is satisfying, given the fact that some shop brands have very similar visual appearance and accuracy of a random guess is below 2\%. After the bit-wise max operation, our style feature achieved an accuracy of 66.14\% with linear regression, as denoted by ``Style+LR'' in Table \ref{table-performance-compare}. 

To better understand what ``style'' exactly represents, we project style features into a 2-D embedding by applying t-SNE\cite{tSNE}, and place each dataset image on a 2-D location according to its style feature. As shown in the colored boxes in Fig. \ref{fig:style-feature-visualize} in left-to-right, top-to-bottom order, we see some interesting patterns within some clusters, like ``red blocks in image'', ``thin doors in image'', ``vertical/horizontal textures'', ``black door plates'' and ``blue door plates''.

\myboldbullet{Text Features}
We tested \cite{SWang:1}'s ngram-score-based method with or without false positive detection (denoted by \textbf{ALL} and \textbf{FPD} in Table \ref{table-performance-compare}), and the shop prediction accuracy is $50.26\%$ and $51.83\%$ respectively. In contrast, our ${\mathcal{G}_k}$ text features with false text detection and a linear classifier (denoted by ``${\mathcal{G}_k}$+LR'') achieved higher accuracy, regardless of the choices of $k \in \{1,2,3,4\}$. This performance improvement is possibly due to the fact that the learned weights rely less on common grams and therefore become more discriminative. Interestingly, ${\mathcal{G}_k}$+LR achieved best accuracy when $k=2$. Our interpretation of this result is that the dictionary of all shop names is usually small and therefore does not require high-order text features to encode. High-order text feature requires a larger number of parameters during model training and therefore could cause overfitting if the training set is small.

\myboldbullet{Fusion} 
As shown in Table \ref{table-performance-compare}, E-FusionNet models outperform style or n-gram classification schemes by a large margin. The best E-FusionNet scheme (using ${\mathcal{G}_2}$ features) achieves the accuracy of 82.55\%, which is consistent with the previous text feature experiment.

\begin{figure}[t]
	\centering
	\includegraphics[width=0.45\textwidth]{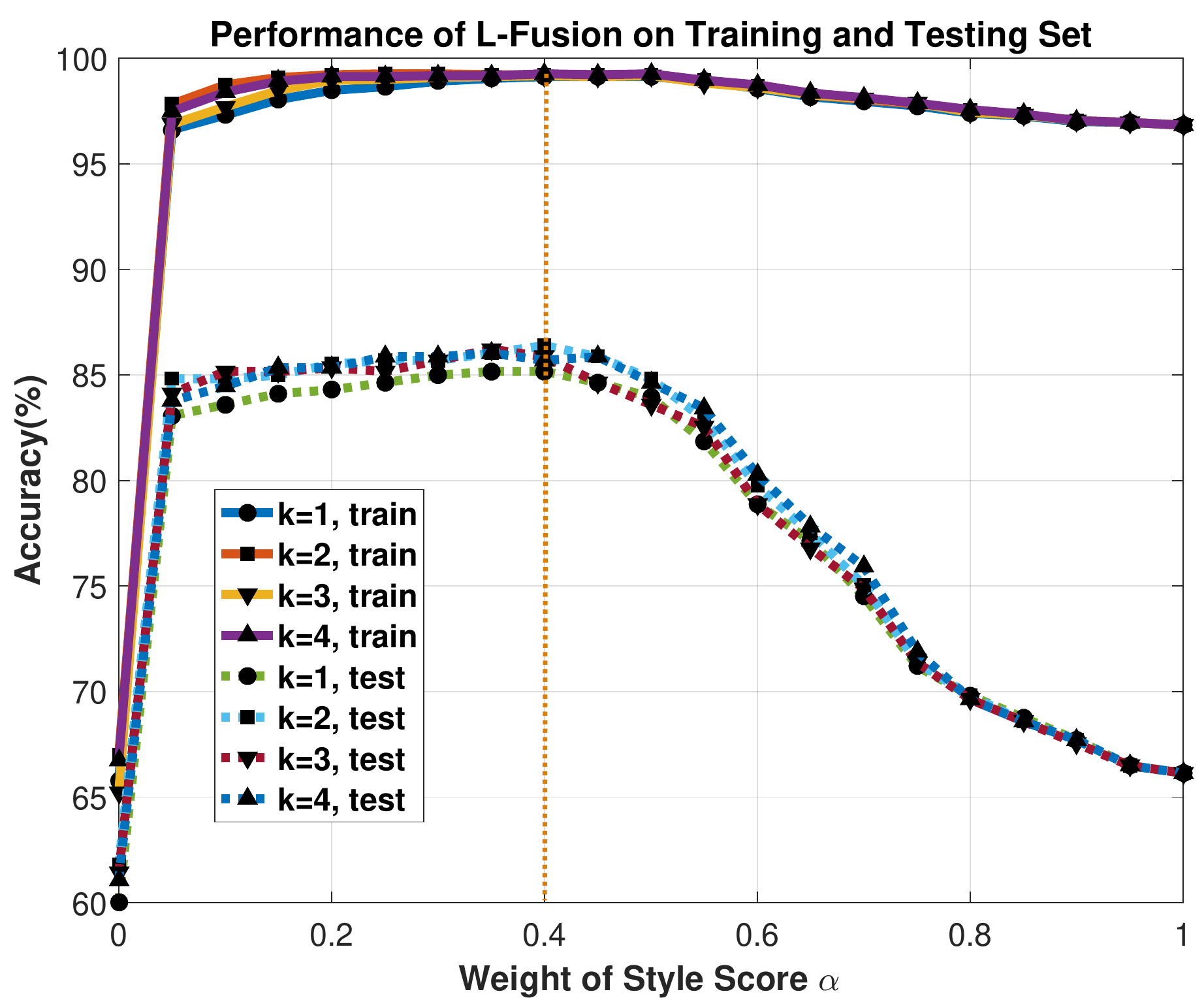}
	\caption{Performance of L-FusionNet model on our dataset under different configurations. $k$ denotes the order of text features.}
	\label{fig-add}
\end{figure}

For L-FusionNet models, we set $\alpha$ in equation \ref{eq-additive-model} to the value which performs best on the training set. All the 4 L-FusionNet models (using ${\mathcal{G}_k}$ with $k\in\{1,2,3,4\}$) perform best on the training set when $\alpha=0.4$ (See Fig. \ref{fig-add}). L-FusionNet model's accuracy at $\alpha=0.4$ is showed in Table \ref{table-performance-compare}. As shown in the table, L-FusionNet reaches the highest accuracy of 86.39\% with ${\mathcal{G}_2}$ features, and outperforms the E-FusionNet model significantly.

Overfitting could be the reason for performance deterioration on E-FusionNet models, because the number of training samples (maximum 2,307) is greatly smaller than the number of parameters required for model training (the feature dimensionality reaches $14096$ when using $\mathcal{G}_4$ features). The L-FusionNet scheme suffers less from this issue, because the dimension of style feature is 4096 and dimension of text feature can be reduced to 558 when using ${\mathcal{G}_2}$.

To see how $\alpha$ affects L-FusionNet models, we further tested L-FusionNet model with different $\alpha$ values on the test set. As shown in Fig. \ref{fig-add}, when $\alpha$ reaches $0$ or $1$, L-FusionNet model degenerates to text-only model or style-only model respectively, and performs poorly. The accuracy rises when $\alpha$ approaches $0.4$ from $0$ or $1$. This observation proves that text features and style features compensate for each other's deficiencies when fused together.

\begin{table}[t]
\centering
\resizebox{\columnwidth}{!}{%
\begin{tabular}{c|cccc|c}
\textbf{Method} & \multicolumn{4}{c}{\textbf{$k$ for ${\mathcal{G}_k}$}} & \textbf{Best} \\ 
 & 1 & 2 & 3 & 4 & \\  
 \hline
 \hline
Style+LR 	& -- & -- & -- & -- & 66.14 \\ 
\hline
Wang \cite{SWang:1} (ALL) 	& -- & -- & -- & -- & 50.26 \\
Wang \cite{SWang:1} (FPD) 	& -- & -- & -- & -- & 51.83 \\
${\mathcal{G}_k}$+LR 	& 60.03 & 61.78 & 61.43 & 61.08 & 61.78 \\ 
\hline
E-FusionNet 	& 78.36 & 82.55 & 81.33 & 80.63 & 82.55 \\
L-FusionNet ($\alpha=0.4$) 	& 85.17 & 86.39 & 85.86 & 85.69 & \textbf{86.39} \\ \hline

\end{tabular} %
}
\caption{Comparison on shop classification accuracy between \cite{SWang:1}'s method, style-only method, text-only method, early FusionNet (E-FusionNet) method and late fusion (L-FusionNet) method on test set.}
\label{table-performance-compare}
\end{table}

\begin{figure}[h!]
\vspace{.3cm}
\begin{minipage}[c]{0.5\textwidth}
\centering
    \includegraphics[width=.75\textwidth]{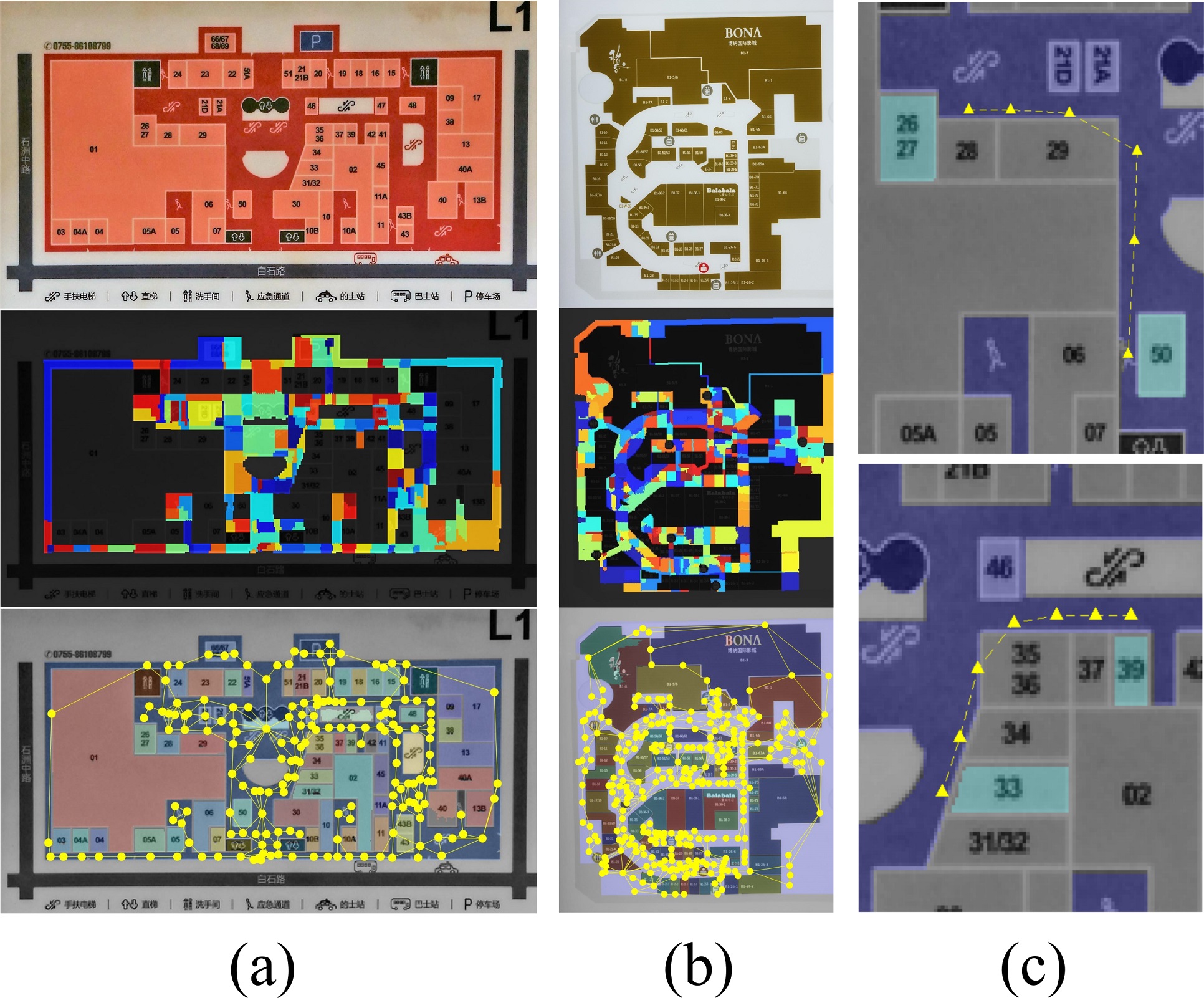}
	\caption{Topological map construction on different indicator maps and simulated path planning. (a)(b) shows two different indicator maps and its segmented road components and the topological maps. (c) shows examples of path planning.}
	\label{topo-navi}
\end{minipage}
\\
\vspace{.3cm}
\\
\begin{minipage}[c]{0.5\textwidth}
	\centering
	\includegraphics[width=.75\textwidth]{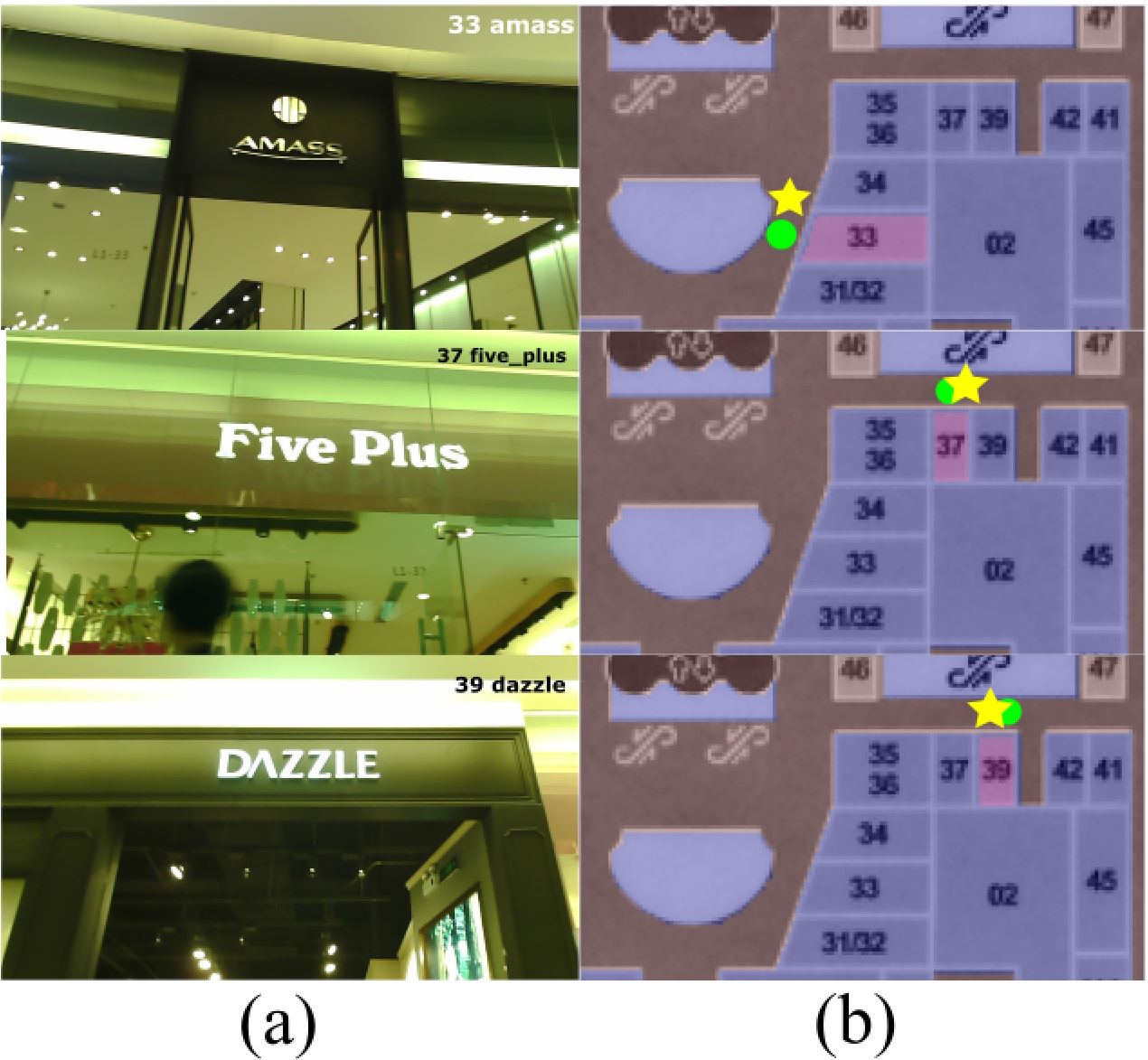}
	\caption{Real-scene localization. Yellow stars indicate averaged position estimation. Green circles indicate the ground truth.}
	\label{localization-result}
\end{minipage}
\end{figure}

\subsection{Map Construction, Navigation and Localization}
To show how our system works in real-life applications, we tested our topological map construction module on shop indicator maps collected from different shops using a cellphone camera. For path planning, a standard Dijkstra algorithm is used on our topological map to find the shortest path between the origin and the destination. For localization, we recorded a video in a shopping mall using a camera installed on a wheelbarrow. A few representative frames containing a storefront are picked out to test our system. Recognized shops are matched with the topological map constructed in Section \ref{section-tp-map-constr} and a small group of nodes is retrieved as the location estimation. 

Experimental results show that our method works on indicator maps with different layouts and in different design styles (see Fig. \ref{topo-navi}). Because information available to the system is limited, in many cases the user will be located on several nodes near the ground truth position. However, if we assume that the user takes photos near each shop and give larger weights to nodes that are closer to the corresponding shop block, a weighted average could be calculated to refine the output (see Fig. \ref{localization-result}).

\section{Conclusion}

In this article, we proposed an indoor positioning system that successfully works in shopping malls. We put forward a feature fusion scheme that fuses high-level style feature and text feature of a storefront image for accurate shop recognition. We designed an automatic method of interpreting shopping instructions for topological map construction. We showed by experiments that feature fusion can improve the accuracy of shop recognition and our system works well in real scenes.

While we see performance improvements by introducing early/late feature fusion, an end-to-end structure could be a better fusion scheme. Also, an abstract topological map may not be precise enough for complicated jobs. However, this problem could be handled by integrating low-level features during SLAM. We leave these questions for future research.

\small
\bibliographystyle{IEEEbib}
\bibliography{bibl}

\end{document}